\documentclass{INTERSPEECH2023}
\pdfoutput 1

\usepackage{booktabs}
\usepackage{multirow}
\usepackage{tabularx}
\usepackage{caption}
\usepackage{hyperref}
\usepackage{svg}
\usepackage{bm}
\usepackage{dcolumn}
\usepackage{siunitx}
\usepackage[justification=centering]{caption}
\newcolumntype{d}[1]{D{.}{.}{#1}}

\interspeechcameraready

\title{The Effects of Input Type and Pronunciation Dictionary Usage\\ in Transfer Learning for Low-Resource Text-to-Speech}
\name{Phat Do$^1$, Matt Coler$^1$, Jelske Dijkstra$^2$, Esther Klabbers$^3$}
\address{
$^1$Campus Fryslân, University of Groningen, the Netherlands\\
$^2$Fryske Akademy/Mercator Research Centre, the Netherlands\\
$^3$ReadSpeaker, the Netherlands}
\email{\{t.p.do, m.coler\}@rug.nl, jdijkstra@fryske-akademy.nl, esther.judd@readspeaker.com}

\begin{document}

\maketitle
 
\begin{abstract}
We compare phone labels and articulatory features as input for cross-lingual transfer learning in text-to-speech (TTS) for low-resource languages (LRLs). Experiments with FastSpeech 2 and the LRL West Frisian show that using articulatory features outperformed using phone labels in both intelligibility and naturalness. For LRLs without pronunciation dictionaries, we propose two novel approaches: a) using a massively multilingual model to convert grapheme-to-phone (G2P) in both training and synthesizing, and b) using a universal phone recognizer to create a makeshift dictionary. Results show that the G2P approach performs largely on par with using a ground-truth dictionary and the phone recognition approach, while performing generally worse, remains a viable option for LRLs less suitable for the G2P approach. Within each approach, using articulatory features as input outperforms using phone labels.

\end{abstract}

\section{Introduction}

\subsection{Multilingual data for TTS in low-resource languages}
\label{intro_1}

Neural TTS produces speech that is both more intelligible and more natural than its preceding paradigms \cite{tanSurveyNeuralSpeech2021}. However, it requires large amounts of training data: LJSpeech \cite{ljspeech17} is one of the most commonly used data sets in neural TTS research and it has nearly 24 hours of professionally recorded single-speaker American English speech. This makes it hard to directly use neural TTS for low-resource languages (LRLs), which make up most of the languages in the world. One approach to deal with this is to make use of data from high-resource languages. This is done by taking advantage of the latent space information shared by languages, even if they are apparently very different. A meta-analysis \cite{doSystematicReviewAnalysis2021b} found this to be effective, and even more so with neural TTS than with earlier TTS paradigms.

In this multilingual data approach, one particular method is cross-lingual transfer learning: pre-training the acoustic model with ample data from a high-resource language (the \textit{``source language''}) and fine-tuning it with limited data in the LRL (the \textit{``target language''}). However, this has an inherent challenge of input mismatch. Two separate languages, however close to each other, almost always have different phone sets. While using grapheme input may avoid the issue of different phone sets, it may also lead to mispronunciation issues. Another challenge is with unseen phones: phones of the target language that are not in the source language. For these, the model has to initialize their associated weights from scratch and learn from the limited data. This makes 
transfer learning less effective. To avoid this, \cite{chenEndtoendTexttospeechLowresource2019} and our previous work \cite{doTexttoSpeechUnderResourcedLanguages2022b} explored solving both challenges simultaneously by applying different methods to map the target languages' phones (\textit{``target phones''}) to their closest counterparts in the source languages (\textit{``source phones''}). By using the pre-trained weights of the source phones, this approach was able to benefit the transfer learning process.

Such phone mapping gives promising results, but it also has issues. First, it may introduce ``accented'' speech. This is when the target phones still (partially) sound like the source phones, especially with very little fine-tuning data. Second, the use of phone labels (e.g., IPA or X-SAMPA symbols) as model input means that the input embeddings are treated in an all-or-nothing manner. In other words, two phones that are represented by two different labels will be considered completely different, even if their pronunciations are close to each other. Consequently, there are likely source phones that are not mapped to any phones and thus simply unused. This leads to inefficient use of data, which is not helpful for LRLs. Both issues may be solved by replacing phone labels (and thus avoiding phone mapping) with universal articulatory features. These are features associated with how the phones are pronounced and can be systematically looked up from the phone labels. The universality of these features also pre-emptively avoids the input mismatch issue, making cross-lingual transfer learning more extensively applicable.

Thus, articulatory features were used as input in transfer learning by~\cite{wellsCrosslingualTransferPhonological2021a}. However, they did not find significant improvements in speech quality compared to using phone labels. Since they used Tacotron~2~\cite{shenNaturalTTSSynthesis2018}, an autoregressive (AR) TTS architecture, their result could be partly due to the architecture's higher data requirement (compared to non-AR TTS~\cite{pineRequirementsMotivationsLowResource2022a}) and its association with unstable attention training~\cite{valentini-botinhaoDetectionAnalysisAttention2021}. Therefore, this study investigates whether using a non-AR TTS architecture can lead to better results in transfer learning with articulatory features.

\subsection{TTS for LRLs without pronunciation dictionaries}
\label{intro_2}

Another issue for LRLs is they often lack a pronunciation dictionary. This limits TTS for such LRLs to two options. The first is using grapheme input, which may lead to mispronunciation issues, which are then compounded in transfer learning. The second is building a pronunciation dictionary or a grapheme-to-phone (G2P) system from scratch. This requires deep linguistic expertise and heavy time investment, both not viable for LRLs.

One approach is to circumvent the need for an explicit dictionary and use G2P conversion during both training and synthesizing. To this end, G2P performance in LRLs can be aided by using multilingual data (similar to TTS acoustic models). One approach that has been used is to train a massively multilingual (269-language) G2P model using found data from Wiktionary, as demonstrated by \cite{liZeroshotLearningGrapheme2022}. To enable parameter sharing across all training languages, they combined data from all languages into a single training set and used special language tokens in the input sequences to distinguish between languages. For predicting each test utterance in an unseen test language, they \textit{1)} searched for the test language's 10 ``nearest'' training languages using a phylogenetic tree, \textit{2)} generated one prediction from each of these 10 languages (conditioned by the corresponding language token) and \textit{3)} ensembled the 10 predictions to form the final prediction. They reached an average phone error rate (PER) of 35.7\% across 605 test languages. Though far from ideal, we argue that this is useful enough for TTS in LRLs
, even more so if articulatory features are used for input, since these can facilitate efficient (transfer) learning even if the phone labels are not identical, as mentioned in Section~\ref{intro_1}.

However, this G2P approach may not work equally well for all LRLs. For any LRL, if its phone inventory and G2P rules are too different from those of its 10 nearest training languages, the approach's performance will likely be lower. It will also degrade if these nearest languages themselves have limited training data and thus the G2P predictions conditioned on them are already not adequately accurate. Therefore, we propose another approach that is theoretically more language-independent: using a universal phone recognizer to predict phone sequences in the audio segments and creating a makeshift pronunciation dictionary using the accompanying texts. This dictionary is then used in the TTS pipeline. This approach is inspired by a recent work in phone recognition by~\cite{liUniversalPhoneRecognition2020}. First, they cascaded a language-independent narrow phone layer and a language-specific broad phone layer in the recognition model. Second, they used phone sets from PHOIBLE~\cite{phoible} (a database of phone inventories of 2,186 languages) to filter the phone output. They reached a PER of 64.2\% for the test language Tusom. While even further from satisfactory than the G2P approach, this approach is more likely to work language-independently. Thus, we posit that this approach can be considered for TTS in LRLs, especially with the use of articulatory features as input. As mentioned in Section~\ref{intro_1}, this input type helps with transfer learning even if the phone labels are not identical and thus is likely more tolerant of inaccuracies in phone prediction.

\subsection{Contributions}
\label{contributions}
Accordingly, we aim to make the following contributions:
\begin{itemize}
    \item[1)]{We use a non-autoregressive architecture (FastSpeech 2~\cite{renFastSpeechFastHighQuality2023}) in cross-lingual transfer learning for TTS in the low-resource language (LRL) West Frisian to compare the effectiveness of phone labels and articulatory features as model input.}
      
    \item[2)]{We explore two options in the absence of a pronunciation dictionary for the target LRL: \textit{a)} predicting phone sequences in both training and synthesizing with a massively multilingual grapheme-to-phone model, and \textit{b)} predicting phone sequences in audio with a universal phone recognizer and building a makeshift pronunciation dictionary.}
    
 \end{itemize}

\section{Data sets \& proposed pipelines}
\subsection{Languages \& data used}
\label{languages_data}
For pre-training, we used the LJSpeech data set to facilitate comparison in future research thanks to its popularity. For the target LRL, we chose West Frisian (``Frysk'', hereafter Frisian), the second official language of the Netherlands with roughly 350,000 native speakers~\cite{gorter2008frisian}. We created a small single-speaker data set from a Frisian audiobook to be used as training data. We split the recordings into utterances by silence and manually checked to obtain their corresponding texts, normalized and expanded the texts where relevant, and trimmed all preceding and trailing silence. We randomly selected 150 utterances ($\sim$15 minutes) for the training and validation set, and another 100 utterances for the test set. Both sets have similar distributions of duration $d$: $\bar{d} = 6.0$, $s_d = 2.2$, $1 \leq d \leq 10$ (seconds). For the baseline models, we used a pronunciation dictionary. This dictionary has roughly 73,000 entries and was derived from the Frisian Audio Mining Enterprise (FAME) project~\cite{yilmazLongitudinalBilingualFrisianDutch2016a}. For the other models, to simulate the lack of pronunciation dictionaries in LRLs, we used this dictionary only to analyze the results.

\subsection{Multilingual grapheme-to-phone (G2P) model}
\label{multilingual_g2p_model}
Following Section~\ref{intro_2}, we used the pre-trained model from~\cite{liZeroshotLearningGrapheme2022} to do G2P conversion for the Frisian data in both training and synthesizing. Using the dictionary as baseline, this model had a mean phone error rate of 33\% $\pm$ 6\% (SE). It should be noted that the model's training data does include Frisian among its 269 training languages, so it is not exactly an unseen language. It is thus reasonable to expect the model to perform better on Frisian than on truly unseen languages. However, the Frisian training data was still rather limited at 991 entries. This places it among more than 95\% of the 874 languages in \cite{liZeroshotLearningGrapheme2022} with fewer than 5,000 entries. We thus assumed that the results from experimenting with Frisian are still relevant for many LRLs, despite technically not being representative for all languages.

\subsection{Universal phone recognition model}
\label{phone_recognition_model}
We used the recognizer checkpoint provided by the authors of~\cite{liUniversalPhoneRecognition2020} to perform phone recognition on the Frisian training and validation sets, limiting the model's output phone pool to the 40-phone Frisian phone set provided by PHOIBLE v2.0. Initial tests showed that the result from this was insufficiently accurate and thus not suitable for TTS training. As a result, we used ground truth phone sequences of 30 utterances ($\sim$3 minutes of speech) to fine-tune the phone recognizer model. Admittedly, such ground truth information is likely not available in practical use with many LRLs, so this partially reduces the approach's applicability for LRLs. We discuss this further in Section~\ref{limitations}.

The fine-tuned model was used to obtain 150 corresponding pairs of texts and predicted phone sequences. Although these were sufficient to train the TTS model, we also needed to generate phone sequences for the test utterances. This is because in a practical scenario, there is no ground truth audio and there is also a need for G2P capability for out-of-vocabulary (OOV) words. There was no straightforward method to do this since we lacked word boundaries information in the audio. Therefore, we first tried using this 150-pair mini data set with OpenNMT v2.0 \cite{kleinOpenNMTOpenSourceToolkit2017} to train a G2P model, but could not reach convergence due to the 
small data size. Consequently, we built a mini makeshift pronunciation dictionary using the predicted phone sequences and their corresponding texts. We manually decided the word boundaries. For words that had more than one predicted phone sequence, we simply chose the most common prediction. We then used this dictionary to train a G2P model so that it could cover all words (including OOVs) in the test set.

\section{Experiment details}
\label{experiment_details}

\subsection{TTS model architecture \& training}
For the acoustic model, we used the FastSpeech 2 implementation ($\sim$35M parameters) of \cite{chienInvestigatingIncorporatingPretrained2021a}. This used phone duration information extracted by the Montreal Forced Aligner (MFA) \cite{mcauliffeMontrealForcedAligner2017}. MFA (v2.0.5) was also used for G2P when needed. For the models with articulatory features as input, we used the implementation in the IMS-Toucan Toolkit v2.3 from \cite{luxLanguageAgnosticMetaLearningLowResource2022}. Phone labels were converted into one-hot encoded vectors of corresponding values from a set of 62 binary articulatory features. This followed the convention in \cite{staibPhonologicalFeatures0shot2020} and complemented their set of 49 features with features such as phone length and more fine-grained values for place and manner of consonant articulation. For the vocoder, we used pre-trained HiFi-GAN V1 ($\sim$14M parameters) \cite{kongHiFiGANGenerativeAdversarial2020b} for all models.

We first pre-trained two models in English, one with phone labels and one with articulatory features as input. Each model was trained for 200,000 parameter updates with a batch size of 32 ($\sim$500 epochs) using the Adam optimizer (\cite{kingmaAdamMethodStochastic2017}, $\beta_1 = 0.9$, $\beta_2 = 0.98$, $\epsilon = 10^{-9}$). We then fine-tuned these two models using the ground-truth Frisian pronunciation dictionary and call the resulting models \textit{ph-gt} and \textit{ft-gt} following their input types. Each fine-tuning was run for 5,000 parameter updates with a batch size of 12 ($\sim$400 epochs). All training was done on a single NVIDIA V100 32GB GPU, taking roughly 45 hours for pre-training and roughly 50 minutes for each fine-tuning.

From the pre-trained models, we also fine-tuned models that did not have access to the dictionary following the pipelines in Section~\ref{multilingual_g2p_model} and~\ref{phone_recognition_model}. We call models from the G2P approach \textit{-g2p} and those from the phone recognition approach \textit{-rec}. To facilitate comparisons between input types, we did this with both phone labels (\textit{ph-}) and articulatory features (\textit{ft-}) input, resulting in four models: \textit{ph-g2p}, \textit{ph-rec}, \textit{ft-g2p}, and \textit{ft-rec}.

\subsection{Evaluation}
\label{evaluation}
We used the fine-tuned models to synthesize the same set of 100 unseen utterances and evaluated them in intelligibility and naturalness. For intelligibility, we used an automatic speech recognition (ASR) model\footnote{\url{huggingface.co/wietsedv/wav2vec2-large-xlsr-53-frisian}} fine-tuned from the ``large'' multilingual checkpoint of the self-supervised learning model \textit{wav2vec 2.0} \cite{baevskiWav2vecFrameworkSelfSupervised2020} on the 50-hour Frisian data set from Common Voice (v8) \cite{ardilaCommonVoiceMassivelyMultilingual2020a}. It reports a word error rate of 16.25\% on Common Voice's test set. We used the ASR model directly (without a language model) on the synthesized utterances and calculated character error rates (CER) for evaluation.

For naturalness, we used an automatic MOS (Mean Opinion Score) prediction model that is based on and fine-tuned from \textit{wav2vec 2.0 Base} \cite{baevskiWav2vecFrameworkSelfSupervised2020}. This follows the approach in \cite{cooperGeneralizationAbilityMOS2022}. Following the result of our work on efficient fine-tuning strategies for MOS prediction in LRLs \cite{doResourceEfficientFineTuningStrategies2023}, we first pre-trained the model on the BVCC data set \cite{cooperGeneralizationAbilityMOS2022} and then further trained it on the neural-TTS-only SOMOS data set \cite{maniatiSOMOSSamsungOpen2022}. Both of these data sets are in English. We then fine-tuned the model further on the MOS data\footnote{\url{phat-do.github.io/sigul22}} from our previous work \cite{doTexttoSpeechUnderResourcedLanguages2022b}. This data contains 2,024 MOS ratings for 220 synthetic utterances (from 11 systems: 10 TTS and 1 resynthesis) rated by 46 participants. The original 100-scale MUSHRA scores were linearly converted to the 5-point MOS scale. We split this data into 80\% training, 10\% validation, and 10\% test sets. Table~\ref{table:mos_pred_accuracy} shows the prediction accuracy of the fine-tuned model on the test set, measured in MSE (Mean Squared Error) and LCC (Linear Correlation Coefficient, i.e., Pearson's \textit{r}). For LCC, corresponding \textit{p}-values were checked for statistical significance. We also include the best-performing measures from the out-of-domain track (OOD) of the recent VoiceMOS Challenge \cite{huangVoiceMOSChallenge20222022}. This OOD track is similar to our settings in terms of cross-lingual MOS prediction with limited ground-truth data. Table~\ref{table:mos_pred_accuracy} shows that the fine-tuned model has an adequate level of prediction accuracy, with an MSE of 0.190 and an LCC of 0.821 for utterance-level predictions. Besides, the test set in this study (described in Section~\ref{languages_data}) has many similarities with the fine-tuning data from \cite{doTexttoSpeechUnderResourcedLanguages2022b}: both are trained with data from audiobooks in Frisian, both share the FastSpeech 2 architecture, etc. In short, we posit that this fine-tuned model was suitable to be used for this study's evaluation and analysis.

\begin{table}[!h]
  \vspace{-1mm}
  \caption{Fine-tuned MOS prediction accuracy}
  \vspace{-3mm}
  \centering
  \begin{tabular}{@{}cccc@{}}
  \toprule
  \textbf{Level} & \textbf{Metric} & \textbf{Current} & \textbf{VoiceMOS's best} \\ \midrule
  \multirow{2}{*}{Utterance}                & MSE &                          $0.190$ & $0.162$ \\
                                            & LCC &                          $0.821$ & $0.921$   \\ \midrule
  \multirow{2}{*}{System}                   & MSE &                       $0.074$    & $0.030$  \\
                                            & LCC &                       $0.946$    & $0.988$       \\
  \bottomrule
  \end{tabular}
  \label{table:mos_pred_accuracy}
  \vspace{-5mm}
  \end{table}

\section{Results \& discussion}
\label{results_discussion}

\begin{figure}[!h]
  \vspace{-2mm}
  \includegraphics[width=\linewidth]{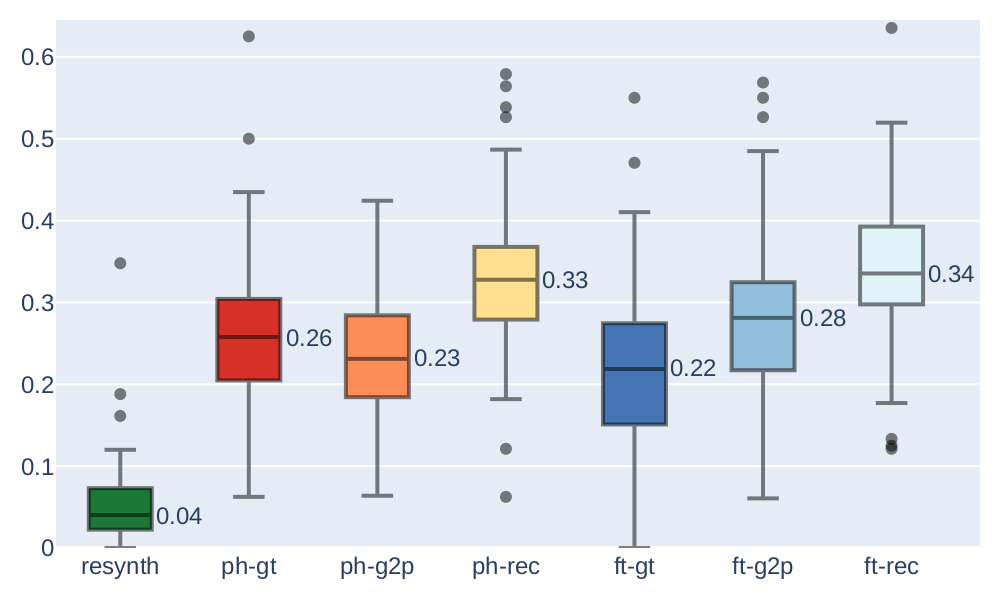}
  \vspace{-8mm}
  \caption{CER on the test set (lower is better)}
  \vspace{-8mm}
  \label{figure_cer}
  \end{figure}

\begin{figure}[!h]
  \includegraphics[width=\linewidth]{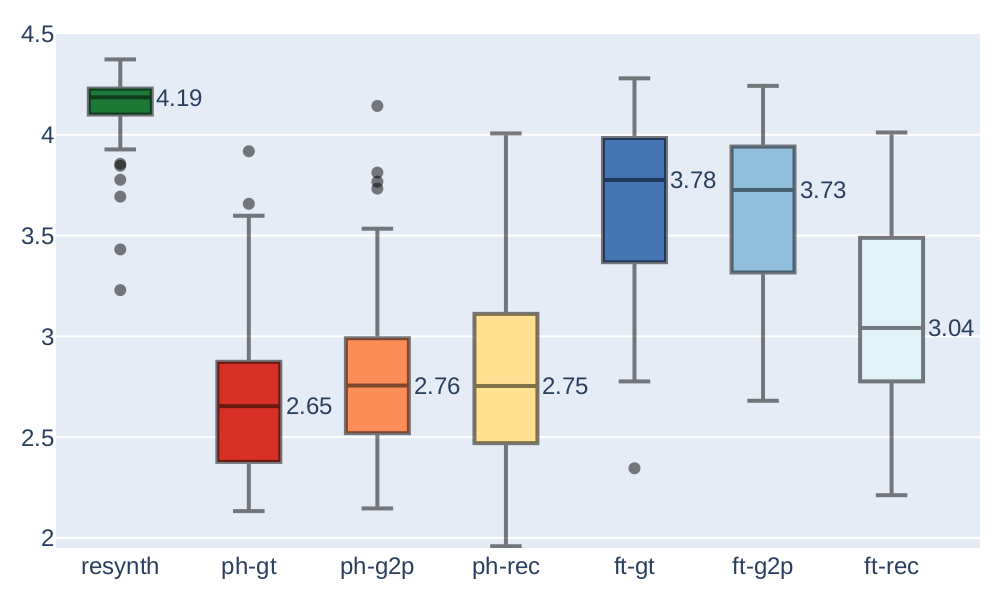}
  \vspace{-8mm}
  \caption{Predicted MOS on the test set (higher is better)}
  \vspace{-2mm}
  \label{figure_mos}
  \end{figure}

We used 100 test utterances for analysis, but for practical reasons, we randomly picked out 20 utterances (140 audio samples) and shared them online for reference\footnote{\url{phat-do.github.io/nodict-IS23}}. Figures~\ref{figure_cer} and~\ref{figure_mos} show the boxplots of CER and MOS for all systems described in Section~\ref{experiment_details}. For reference, we also included results from resynthesized audio (generated from ground-truth spectrograms).

We used a linear mixed effects model and treated input type (\textit{ph} and \textit{ft}) and ``dictionary type'' (\textit{gt}, \textit{g2p}, and \textit{rec}), plus their interaction, as fixed effects. For random effects, we used random intercepts for utterances and by-utterance random slopes for both input type and dictionary type. Residual plots were used to check for assumptions of homoscedasticity and normality. Statistical significance was checked using \textit{p}-values from the likelihood ratio tests between models with and without each effect in question. The results from this show that input type, dictionary type, and their interaction all had significant effects on both CER and MOS. Therefore, we conducted a post-hoc analysis using Tukey's multiple comparison of means \cite{haynesTukeyTest2013} to check if the mean CER and MOS of all scenarios (combinations of input and dictionary type) were significantly different from each other. To avoid cluttering, only relevant scenarios are included in Table~\ref{table:tukey}. Significant differences are in bold.

\begin{table}[!h]
  \centering
  \vspace{-2mm}
  \caption{Results from Tukey's multiple comparison of means}
  \vspace{-2mm}
  \label{table:tukey}
  \addtolength{\tabcolsep}{-0.38em}
  \begin{tabular}{@{}cc|cc|rr|rr@{}}
  \toprule
  \multicolumn{2}{c}{\textbf{Group 1}} &  \multicolumn{2}{c}{\textbf{Group 2}} & \multicolumn{4}{c}{$\mathbf{\mu_{G2} - \mu_{G1}}$ \& \textit{p}-value} \\ \midrule
  Input   & Dict   & Input   & Dict   &  \multicolumn{2}{c}{\textbf{CER}}  & \multicolumn{2}{c}{\textbf{MOS (predicted)}} \\ \midrule
  
  {\textit{ph}}         & \textit{gt}   & {\textit{ft}}         & \textit{gt}   & $\mathbf{-0.045}$                            & $.003$                     & $\mathbf{0.982}$                           & $<.001$           \\
  \midrule

  \multirow{3}{*}{\textit{ph}}         & \textit{gt}   & \multirow{3}{*}{\textit{ph}}         & \textit{g2p}    & $-0.026$                                    & $.267$                              & $0.101$                                     & $.386$                              \\
                                        & \textit{gt}   &                             & \textit{rec}    & $\mathbf{0.065}$                            & $<.001$           & $0.114$                                     & $.341$                              \\
                                      & \textit{g2p}    &                             & \textit{rec}    & $\mathbf{0.092}$                           & $<.001$           & $0.004$                                    & $.999$                                  \\ \midrule
  \multirow{3}{*}{\textit{ft}}         & \textit{gt}   & \multirow{3}{*}{\textit{ft}}         & \textit{g2p}    & $\mathbf{0.055}$                            & $<.001$           & $-0.021$                                     & $.999$                              \\
                                      & \textit{gt}   &                             & \textit{rec}    & $\mathbf{0.125}$                            & $<.001$           & $\mathbf{-0.543}$                           & $<.001$           \\
                                      & \textit{g2p}    &                             & \textit{rec}    & $\mathbf{0.070}$                           & $<.001$           & $\mathbf{-0.522}$                            & $<.001$             \\ \midrule
  \multirow{2}{*}{\textit{ph}}         & \textit{g2p}   & \multirow{2}{*}{\textit{ft}}         & \textit{g2p}   & $\mathbf{0.035}$                            & $.042$                     & $\mathbf{0.851}$                           & $<.001$           \\ 
  & \textit{rec}    &                             & \textit{rec}    & $0.014$                           & $.857$                     & $\mathbf{0.324}$                           & $<.001$           \\
   
                                      \bottomrule

  \end{tabular}
  \vspace{-5mm}            
  \end{table}

\subsection{Input type comparison with ground-truth dictionary}
With \textit{gt}, \textit{ft} decreased CER by $0.045$ ($\pm0.01$) and increased MOS by $0.982$ ($\pm0.05$) compared to \textit{ph}. In other words, with all things equal (the same TTS architecture, training data, training procedure, evaluation method, etc.), using articulatory features as input outperformed using phone labels for cross-lingual transfer learning. This likely comes from the better learning efficiency hypothesized in Section~\ref{intro_1}. As a result, TTS for LRLs is expected to benefit from this. This is at least with a non-autoregressive (non-AR) architecture as used in this study, as opposed to the AR architecture in \cite{wellsCrosslingualTransferPhonological2021a}.

\subsection{Multilingual G2P \& phone recognition}
\textbf{Phone labels input:} Compared to \textit{gt}, \textit{g2p} did not significantly change either CER or MOS. Meanwhile, \textit{rec} increased CER by $0.065$ ($\pm0.01$) compared to \textit{gt} and $0.092$ ($\pm0.01$) compared to \textit{g2p}. This means that with phone labels, using multilingual G2P was a viable approach, giving output speech quality comparable to that from the ground-truth dictionary. The phone recognition approach was less effective, producing a similar level of naturalness but worse intelligibility.

\textbf{Articulatory features input:} Compared to \textit{gt}, \textit{g2p} led to an increase of $0.055$ ($\pm0.01$) in CER, but no significant change in MOS. However, \textit{rec} increased CER by $0.125$ ($\pm0.01$) and decreased MOS by $0.543$ ($\pm0.04$) compared to \textit{gt}. This means that with articulatory features, multilingual G2P was only comparable to the ground-truth dictionary in naturalness, and phone recognition led to worse quality in both criteria.

\textbf{Phone labels vs. articulatory features:} For \textit{g2p}, \textit{ft} led to both an increase of $0.035$ ($\pm0.006$) in CER and an increase of $0.851$ ($\pm0.03$) in MOS. For \textit{rec}, \textit{ft} had no effect on CER, but increased MOS by $0.324$ ($\pm0.03$). In other words, in the absence of an actual pronunciation dictionary, articulatory features outperformed phone labels in naturalness while performing either worse or similarly in intelligibility. This is partially in line with our hypothesis in Section~\ref{intro_2}.

\subsection{Assumptions \& limitations}
\label{assumptions}
\textbf{Assumptions:} We assumed that the use of CER calculated from ASR predictions was relevant in evaluating the intelligibility of synthetic speech, i.e., it is comparable to human evaluation. This is especially notable since we used an ASR model without a language model. Meanwhile, humans cannot really turn off their language models \cite{lakhotiaGenerativeSpokenLanguage2021a}. For MOS evaluation, besides the usual assumption that MOS can be reliably used to judge ``naturalness'', we assumed that our model was fine-tuned enough to give scores representative of those made by human listeners.

\label{limitations}
\textbf{Limitations:} As mentioned in Section~\ref{multilingual_g2p_model}, Frisian was among the G2P model's training data (albeit with limited data). Therefore, we could not reliably infer that the approach would work equally well for truly unseen LRLs. For the phone recognition model, since it relies on phone sets from PHOIBLE, its performance would degrade for an LRL not among the 2,186 languages covered by this database. Even for a covered language, the phone recognition model required some fine-tuning before being directly integrated into the TTS pipeline due to its base performance. This fine-tuning may require some expertise in phonetics and/or the LRL, which may limit its applicability.

\section{Conclusions \& future work}
In this study, we explored cross-lingual transfer learning from English (LJSpeech) to West Frisian, using 15 minutes of audiobook data. The output speech was evaluated with CER obtained from an ASR model and MOS from an MOS prediction model, both based on \textit{wav2vec 2.0}. The results confirmed that using articulatory features as input for the TTS acoustic model significantly outperformed using phone labels in both intelligibility and naturalness. Accordingly, we believe that this approach is  beneficial for similar work in low-resource languages (LRLs).

For LRLs without available pronunciation dictionaries, we propose two approaches. The first is doing grapheme-to-phone (G2P) conversion directly in both training and synthesizing with a massively multilingual G2P model. The second is creating a makeshift dictionary using a universal phone recognition model. The results of our experiments indicated that the G2P approach was generally comparable to a ground-truth dictionary, while the phone recognition approach only showed comparable results in one case. Thus, G2P is a viable option in almost all cases while phone recognition is more limited but remains an option for LRLs that cannot benefit much from G2P. The two approaches' effects also differed between the input types: articulatory features generally led to better speech quality than phone labels. These findings are expected to enable and provide strategies toward developing TTS for the thousands of LRLs that have no available pronunciation dictionaries.

Future research is planned to explore bypassing the intermediate step of phonetic transcriptions (to skip a potential source of inaccuracy) and converting audio directly to articulatory features to be used in creating a makeshift dictionary.

\section{Acknowledgements}
We thank the Center for Information Technology of the University of Groningen for providing access to the Hábrók high performance computing cluster. We are thankful for the permission to use the recordings by Geartsje de Vries of books written by Koos Tiemersma and published by the publisher Audiofrysk.

\bibliographystyle{IEEEtran}
\bibliography{mybib}

\end{document}